# Enriching WordNet concepts with topic signatures

**Eneko Agirre, Olatz Ansa, David Martinez**
IXA NLP Group
University of the Basque Country
649 pk. 20.080
Donostia. Spain
eneko@si.ehu.es

**Eduard Hovy**
USC Information Sciences Institute
4676 Admiralty Way
Marina del Rey, CA 90292-6695
hovy@isi.edu

**Abstract**

This paper explores the possibility of enriching the content of existing ontologies. The overall goal is to overcome the lack of topical links among concepts in WordNet. Each concept is to be associated to a topic signature, i.e., a set of related words with associated weights. The signatures can be automatically constructed from the WWW or from sense-tagged corpora. Both approaches are compared and evaluated on a word sense disambiguation task. The results show that it is possible to construct clean signatures from the WWW using some filtering techniques.

## Introduction

Knowledge acquisition is a long-standing problem in both Artificial Intelligence and Computational Linguistics. Semantic. Huge efforts and investments have been made to build repositories with such knowledge (which we shall call ontologies for simplicity) but with unclear results, e.g. CYC (Lenat, 1995), EDR (Yokoi, 1995), WordNet (Fellbaum, 1998). WordNet, for instance, has been criticized for its lack of relations between related concepts.

As an alternative to entirely hand-made repositories, automatic or semi-automatic means have been proposed for the last 30 years. On the one hand, shallow techniques are used to enrich existing ontologies (Hearst & Schutze, 1993) or to induce hierarchies (Caraballo, 1999), usually analyzing large corpora of texts. On the other hand, deep natural language processing is called for to acquire knowledge from more specialized texts (dictionaries, encyclopedias or domain specific texts) (Wilks, 1996; Harabagiu et al. 1999). These research lines are complementary; deep understanding would provide specific relations among concepts, whereas shallow techniques could provide generic knowledge about the concepts.

(Agirre et al. 2000) explores preliminary experiments in exploiting text on the world wide web in order to enrich WordNet. Each concept in WordNet was linked to relevant document collections in the web. These document collections were processed in order to link the concepts to topically related words, which form the *topic signature* for each concept in the hierarchy (see Table 1 for an example of topic signatures). Although promising, the signatures contained noisy terms, which could be eliminated with further work. Two improvement areas were mentioned in particular: the method of linking concepts to documents in the web and the way of computing the weights of the signatures.

Evaluation of automatically acquired semantic and world knowledge information is not an easy task. There is no gold standard for topic signatures and hand evaluation is arbitrary. In this paper we used manual comparison of the signatures obtained from the WWW with those obtained from two sense-tagged corpora, the DSO corpus (Ng and Lee, 1996) and Semcor (Miller et al. 1993), shedding some light in the nature of this kind of extracted information. The extracted signatures are on the web for inspection (Agirre et al. 2001). The other evaluation criteria has been word sense disambiguation (WSD): topic signatures were used to tag a given occurrence of a word with the intended concept. The benchmark corpus for evaluation was SemCor. Our aim is not to compete with other word sense disambiguation algorithms, but to test whether the acquired knowledge is valid.

The structure of the paper is as follows. The first section reviews WordNet and the benchmark corpora. Section 2 presents the basic method of building the topic signatures from the WWW and from the sense-tagged corpora. Some possible improvements are presented in Section 3, with comments on the quality of the signatures. Section 4 presents the results on a word sense disambiguation task. Finally some conclusions are drawn and further work is outlined.

## 1 Brief introduction to WordNet and the sense-tagged corpora.

WordNet is an online lexicon based on psycholinguistic theories (Fellbaum,1998). It comprises nouns, verbs, adjectives and adverbs, organized in terms of their meanings around lexical-semantic relations, which include among others, synonymy and antonymy, hypernymy and hyponymy (similar to *is-a* links), meronymy and holonymy (similar to *part-of* links). Lexicalized concepts, represented as sets of synonyms called synsets, are the basic elements of WordNet. The version used in this work, WordNet 1.6, contains 121,962 words and 99,642 concepts.

The noun *church*, for instance, has 3 word senses (in other words lexicalized concepts or word-concept pairs). For the sake of simplicity we use *concept* to refer to both word sense and concept itself. The set of synonyms and gloss for each sense of church is shown in Figure 1.

Being one of the most commonly used semantic resources in natural language processing, some of its shortcomings are broadly acknowledged, e.g., the lack of links between topically related concepts: there is no link between pairs like *bat–baseball, fork–dinner, farm–chicken,* etc. The signatures presented in this paper are a first step in providing such links.

SemCor (Miller et al. 1993) is a corpus in which word sense tags (which correspond to WordNet concepts) have been manually included for all open-class words in a 360,000-word subset of the Brown Corpus (Francis and Kucera, 1989). We use SemCor to evaluate the topic signatures in a word sense disambiguation task. In order to choose a few nouns to perform our experiments, we focused on a random set of 7 nouns[1].

DSO (Ng and Lee, 1996) comprises around 1000 sentences for 191 nouns and verbs, where only the target word was disambiguated. The sentences are taken from the Brown corpus and the Wall Street Journal (WSJ). We selected the documents coming from the WSJ only, as the brown corpus is already covered in Semcor.

## 2 Building topic signatures for the concepts in wordnet

In this work we want to collect for each concept in WordNet the words that appear most distinctively in texts related to it. That is, we aim at constructing lists of closely related words for each concept. For example, WordNet provides two possible word senses or concepts for the noun *waiter*:
*1: waiter, server — a person whose occupation is to serve at table (as in a restaurant)*
*2: waiter — a person who waits or awaits*

For each of these concepts we would expect to obtain two lists with words like the following:
*1: restaurant, menu, waitress, dinner, lunch, etc.*
*2: hospital, station, airport, boyfriend, girlfriend, etc.*

We can build such lists from a sense-tagged corpora just observing which words co-occur distinctively with each sense, or we can try to associate a number of documents from the WWW to each sense and then analyze the occurrences of words in such documents.

The method is organized as follows. We first organize the documents in collections, one collection per word sense: directly using sense-tagged corpora, or exploiting the information in WordNet to build queries, which are used to search in the Internet those texts related to the given word sense. Either way we get one document collection per word sense. For each collection we extract the words and their frequencies, and compare them with the data in the other collections. The words that have a distinctive frequency for one of the collections are collected in a list, which constitutes the topic signature for each word sense. The steps are further explained below.

### 2.1 Building the queries

The original goal is to retrieve from the web all documents related to an ontology concept. If we assume that such documents have to contain the words that lexicalize the concept, the task can be reduced to classifying all documents where a given word occurs into a number of collections of documents, one collection per word sense. If a document cannot be classified, it would be assigned to an additional collection.

The goal as phrased above is unattainable, because of the huge amount of documents involved. Most of words get millions of hits: *boy* would involve retrieving 2,325,355 documents, *church* 6,243,775, etc. Perhaps in the future a more ambitious approach could be tried, but at present we cannot aim at classifying those enormous collections. Instead, we construct queries, one per concept, which are fed to a search engine. Each query will retrieve the documents related to that concept.

The queries are constructed using the information in the ontology. We followed the method of (Mihalcea and Moldovan, 1999). Four procedures are defined to query the search engine in order: use monosemous synonyms, use the defining phrases, use synonyms with the AND operator and words from the defining phrase with the NEAR operator, and lastly, use synonyms and words from the defining phrases with the AND operator. The procedures are sorted by preference, and one procedure is only applied if the previous one fails to retrieve any examples. We fed the queries into altavista (AltaVista, 2001), retrieving 150 documents per word sense.

---

[1] The reason for using only 7 words is that retrieving documents from the web was slower than foreseen.

*1: church, Christian church, Christianity - a group of Christians; any group professing Christian doctrine or belief*
*2: church, church building - for public (especially Christian) worship*
*3: church service, church - a service conducted in a church*

**Figure 1**. Synsets and glosses for the three senses of church.

| Semcor full documents | | | | | | WWW sentences | | | | | |
|---|---|---|---|---|---|---|---|---|---|---|---|
| *Church1* | | *Church2* | | *Church3* | | *Church1* | | *Church2* | | *Church3* | |
| **catholic** | 59.03 | n't | 18.38 | **mike** | 128.64 | together | 18.4 | new | 17.6 | **sunday** | 20.5 |
| **spirit** | 43.97 | mr. | 13.59 | **jury** | 122.21 | christ | 12.6 | **old** | 15.3 | **attend** | 12.6 |
| **protestant** | 42.95 | do | 12.08 | 1 | 102.52 | **god** | 9.00 | construction | 11.4 | yet | 9.8 |
| **rector** | 32.72 | out | 11.94 | election | 96.48 | biblical | 6.59 | **congregation** | 10.2 | special | 8.4 |
| **congregation** | 32.72 | door | 10.90 | **fulton** | 90.05 | jesus | 6.44 | build | 8.9 | pastor | 7.0 |
| **petitioner** | 30.68 | **wilson** | 9.41 | sam | 90.05 | **christian** | 5.91 | building | 8.5 | t | 7.0 |
| **community** | 27.61 | window | 8.92 | **county** | 82.91 | believe | 5.27 | committee | 7.6 | a.m. | 7.0 |
| **economic** | 27.49 | river | 8.92 | **mars** | 57.89 | gather | 5.27 | house | 6.3 | child | 5.7 |
| **report** | 27.09 | **house** | 8.54 | **resolution** | 57.89 | just | 5.27 | erect | 6.3 | **morning** | 5.6 |
| **christian** | 26.59 | wall | 8.42 | funds | 51.45 | gospel | 5.27 | plan | 6.3 | air | 5.6 |
| **mission** | 26.59 | papa | 7.93 | **georgia** | 51.45 | friend | 5.27 | **episcopal** | 6.3 | breakfast | 5.6 |
| **hearing** | 25.56 | big | 7.77 | bond | 51.45 | strive | 5.27 | project | 6.3 | night | 5.6 |
| **england** | 25.56 | **bed** | 7.77 | 2 | 45.52 | share | 5.27 | corner | 5.1 | include | 5.6 |
| **social** | 24.45 | **street** | 7.47 | **candidate** | 45.02 | **roman** | 5.27 | avenue | 5.1 | day | 5.6 |
| **board** | 22.53 | **mouth** | 7.43 | **martian** | 45.02 | truth | 5.27 | complete | 5.1 | sing | 5.6 |
| **john** | 21.87 | **look** | 6.91 | highway | 45.02 | say | 5.25 | construct | 5.1 | please | 5.6 |
| claim | 21.47 | **old** | 6.65 | **atlanta** | 45.02 | small | 3.96 | design | 5.1 | **prayer** | 4.2 |
| **roman** | 21.47 | talk | 6.61 | vote | 45.02 | hear | 3.95 | road | 5.1 | 10 | 4.2 |
| officer | 20.45 | think | 6.45 | petition | 38.59 | live | 3.95 | **continue** | 5.1 | **preach** | 4.2 |
| such | 18.99 | car | 6.44 | campaign | 38.59 | loving | 3.95 | state | 5.1 | next | 4.2 |
| position | 18.41 | get | 6.32 | legislator | 38.59 | business | 3.95 | work | 3.8 | **class** | 4.2 |
| **michelangelo** | 18.41 | come | 5.99 | republican | 32.16 | **faith** | 3.95 | white | 3.8 | monthly | 4.2 |

**Table 1**. Topic signatures for church. On the left the signatures created using full documents from the sense-tagged Semcor. On the right the signatures created using the sentence context from documents extracted from the WWW. On bold, terms which are left after filtering with the word signature for church with $\chi^2$ values over 4.64.

## 2.2 Build topic signatures

The document collections retrieved in step 2.1 are used to build the topic signatures. The internet documents are processed in order to extract textual words, and their frequencies in the document collection. We thus obtain one vector for each collection, that is, one vector for each word sense of the target word.

In order to measure which words appear distinctively in one collection in respect to the others, a signature function was selected based on previous experiments (Lin, 1997; Hovy and Junk, 1998). We needed a function that would give high values for terms that appear more frequently than expected in a given collection. The signature function that we used is $\chi^2$, which we will define next.

The vector $vf_i$ contains all the words and their frequencies in the document collection $i$, and is constituted by pairs $(word_j, freq_{i,j})$, that is, one word $j$ and the frequency of the word $j$ in the document collection $i$. We want to construct another vector $vx_i$ with pairs $(word_j, w_{i,j})$ where $w_{i,j}$ is the $\chi^2$ value for the word $j$ in the document collection $i$ (cf. Equation 1).

$$w_{i,j} = \begin{cases} \frac{(freq_{i,j} - m_{i,j})}{m_{i,j}} & \text{if } freq_{i,j} > m_{i,j} \\ 0 & \text{otherwise} \end{cases} \quad (1)$$

Equation 2 defines $m_{i,j}$, the expected mean of word $j$ in document $i$.

$$m_{i,j} = \frac{\Sigma_j freq_{i,j} \; \Sigma_j freq_{i,j}}{\Sigma_{i,j} freq_{i,j}} \quad (2)$$

When computing the $\chi^2$ values, the frequencies in the target document collection are compared with the rest of the document collection, which we call the *contrast set*. In this case the contrast set is formed by the other word senses. Table 1 shows some signatures for church.

## 3 Filtering the signatures.

Hand inspection of the signatures constructed following the method above showed that some terms with high weight were strange. For instance:
  Boy: anything.com, xena, tpd-results, etc.
  Church: ai, ruby, le, lee, etc.

The following strategies were tried to filter the signatures:
1. Only take one document from each website.
2. Lemmatize the source documents and keep only open category lemmas
3. Limit the context to the sentence
4. Build a signature for the target word using a reference corpus and filter the concept related signatures keeping only the most relevant terms[2] in the general signature.

The first two improved greatly the visual quality of the signatures retrieved, as most of non-sensical terms disappeared. But there were still some disturbing terms in the signatures. Table 1 shows the topic signatures for *church*[3], as created using the sense tags in Semcor after applying 1 and 2. We observed that some specific and rare terms (cf. *mike, sam, fulton*, etc. in Table 1) which appeared often in a single document could happen to be included in the signature.

In order to limit the effect of such words we tried the following alternatives: to limit the context to just the surrounding sentence (3) and to limit the possible terms to those that co-occur often with the target word (4). The later was implemented constructing a signature for the target word, using as reference corpus the whole brown corpus, comparing the documents that contained the target word against the documents where the target word did not appear. A cutoff value of 4.64 was used, giving a confidence value of 0.90. Strategies 3 and 4 could happen to discard interesting terms so we experimented with both in the WSD exercise.

Table 1 shows on the right the topic signature for church constructed from the WWW using only sentences (3). These signatures do not have rare terms like those from full documents. The effect of applying the word signature (4) is shown also in Table 1, as terms which are kept after filtering are shown in bold. The filter has no much effect on the Semcor signature, but it helps select terms from the WWW signature.

The topic signatures for church from the web are better than those coming from Semcor, specially for senses with lower frequency in Semcor like *church2* and *church3*.

## 4 Evaluating the signatures on a WSD task

The goal of this experiment is to evaluate the automatically constructed topic signatures, not to compete against other word sense disambiguation algorithms. If topic signatures yield good results in word sense disambiguation, it would mean that topic signatures have correct information, **and** that they are useful for word sense disambiguation. Topic signatures constitute one source of evidence (possibly weak), but do not replace other sources (local collocations, etc). Therefore, we do not expect impressive results.

Given the following sentence from SemCor, a word sense disambiguation algorithm should decide that the intended meaning for *waiter* is that of a restaurant employee:
  "... and threatened to shoot a <u>waiter</u> who was pestering him for a tip."

The word sense disambiguation algorithm is straightforward. Given an occurrence of the target word in the text we collect the words in its context, and for each word sense we retrieve the $\chi^2$ values for the context words in the corresponding topic signature. For each word sense we add these $\chi^2$ values, and then select the word sense with the highest value. Different context sizes have been tested in the literature, and large windows have proved to be useful for topical word sense disambiguation. We chose a window of 100 words around the target for signatures constructed with full documents and just the sentence for sentence signatures.

In order to compare the results, we computed a number of baselines. First of all choosing the sense at random (ran). We also constructed lists of related words using WordNet, in order to compare their performance with that of the signatures: the list of synonyms (Syn), these plus the content words in the definitions (S+def), and these plus the hyponyms, hypernyms and meronyms (S+all). The algorithm to use these lists is the same as for the topic signatures. The signatures constructed from the web are compared to other signatures constructed from manually sense-tagged corpora: Semcor and WSJ (cf. section 2). The results for Semcor constitute an upper bound, as they are constructed from the same data they are tested. The two filtering strategies are also evaluated separately: sentence context and filtering using word vector[4] (cf. section 3).

---

[2] As given by a $\chi^2$ cutoff value.

[3] The signatures for all words can be checked at (Agirre et al. 2001).

[4] Signatures from Semcor where not filtered, as the word

|         |    |      | WWW    | Baselines |      |       |       | Doc    |      | Sent   |      |      | Filter Doc | Filter Sent |      |
|---------|----|------|--------|-----------|------|-------|-------|--------|------|--------|------|------|------------|-------------|------|
| word    | #s | #occ | method | Ran       | Syn  | S+def | S+all | semcor | www  | semcor | wsj  | www  | www        | wsj         | www  |
| Account | 8  | 27   | 2.02   | 0.12      | 0.09 | 0.28  | 0.20  | 0.74   | 0.30 | 0.59   | -    | 0.19 | 0.26       | -           | 0.14 |
| Age     | 4  | 104  | 2.02   | 0.25      | 0.00 | 0.04  | 0.03  | 0.65   | 0.22 | 0.43   | 0.24 | 0.41 | 0.16       | 0.22        | 0.51 |
| Boy     | 4  | 169  | 3.19   | 0.25      | 0.48 | 0.37  | 0.59  | 0.13   | 0.30 | 0.61   | -    | 0.31 | 0.30       | -           | 0.36 |
| Church  | 3  | 128  | 1.31   | 0.33      | 0.28 | 0.49  | 0.46  | 0.66   | 0.46 | 0.76   | 0.28 | 0.45 | 0.48       | 0.20        | 0.40 |
| Duty    | 3  | 25   | 3.10   | 0.33      | 0.08 | 0.47  | 0.58  | 0.88   | 0.44 | 0.68   | -    | 0.52 | 0.44       | -           | 0.52 |
| Interest| 7  | 139  | 2.57   | 0.14      | 0.04 | 0.03  | 0.07  | 0.47   | 0.18 | 0.68   | 0.33 | 0.15 | 0.13       | 0.28        | 0.12 |
| Member  | 4  | 74   | 3.34   | 0.25      | 0.01 | 0.76  | 0.61  | 0.08   | 0.05 | 0.62   | 0.15 | 0.37 | 0.11       | 0.19        | 0.22 |
| Total   |    | 666  |        | 0.24      | 0.19 | 0.31  | 0.36  | 0.43   | 0.27 | 0.63   | -    | 0.33 | 0.26       | -           | 0,30 |
| Total WSJ |  | 445  |        | 0.24      | 0.10 | 0.29  | 0.26  | 0.50   | 0.25 | 0.64   | 0.27 | 0.34 | 0.23       | 0,23        | 0,31 |

**Table 2**. Word sense disambiguation results given as recall. *Total* shows average results for all words, and *Total WSJ* for the words in the WSJ corpus.

Table 2 shows the results for the selected nouns. The number of senses attested in SemCor (#s) and the number of occurrences of the word in SemCor (#occ) are also presented. The results are given as recall, that is, the number of successful tags divided by the total number of occurrences. Multiple answers are penalized, simulating random choice. The results show the following:

- WWW signatures are well above random baseline, and above Wordnet baselines for most of the words, showing that useful information not present in WordNet has been learned.
- Semcor signatures do constitute an upper bound, as expected: testing is performed on the same texts from which the signatures were constructed.
- Signatures from WSJ perform badly, perhaps because coming from a single domain they have been applied to a balanced corpus.
- Limiting the context for extraction to the sentence improves the performance for both Semcor and WWW signatures. This proves to be a useful way to clean the signatures.

## 5 Discussion and comparison with related work

Both manual inspection and word sense disambiguation results show that the automatically constructed signatures learn words that are topically related to the target concepts. The quality of the signatures changes from one word to the other, specially in the case of word senses that are closely related. This evidence was used successfully to cluster word senses in (Agirre et al. 2000). Nevertheless, it is clear from the word sense disambiguation results that signatures are not that useful for word sense disambiguation.

According to manual inspection (cf. Table 1), filtering and sentence context are useful to get clean signatures. The results on the word sense disambiguation task, though, show that for some cases some terms which were useful for disambiguating were lost.

Next we review some related work.

### 5.1 Building topic signatures

All references build topic signatures for general *topics* rather than for concepts in an ontology. Topic signatures were an extension of relevancy signatures (Riloff, 1996) developed for text summarization (Lin, 1997). Lin's topic signature construction is similar to ours, except that he used *tf.idf* for term weighting. In subsequent work, Hovy and Junk (1998) explored several alternative weighting schemes, finding that $\chi^2$ provided better results than *tf.idf* or *tf*. Lin and Hovy (2000) use a likelihood ratio from maximum likelihood estimates that achieves even better performance on clean data.

### 5.2 Searching the internet for concepts

(Mihalcea and Moldovan, 1999) present the method used in this paper. They evaluate the results by hand with very good results. (Agirre et al. 2000) used a method which tried to constrain as little as possible the cues used to build the queries. As a result poorer signatures were constructed. In general, documents retrieved from the web introduce a certain amount of noise into signatures, which can be alleviated with the methods introduced in section 3.

### 5.3 Uses of topic signatures for ontology enrichment.

(Junk and Hovy 1998) explore the integration of topic signatures in Sensus. Their results indicate that further attention is required if the hierarchical structure is to be profited. (Agirre et al. 2000)

---

signature comes from the very same source.

presents preliminary results in word sense clustering using similarity among topic signatures, which could help alleviate the word sense proliferation in WordNet.

## 6 Conclusions and further research

We have introduced an automatic method to enrich automatically the concepts in WordNet with topic signatures, using the huge amount of documents in the world wide web. Both manual inspection and word sense disambiguation results show that the quality of the signatures acquired is comparable to those from a balanced hand-tagged corpus, and better than those from a specialized hand-tagged corpus. The filtering techniques have been useful on getting clean signatures. All extracted signatures can be checked at (Agirre et al. 2001).

We have to stress that topic signatures have not been constructed with word sense disambiguation in mind. In that respect, it is not surprising that the disambiguation results are not spectacular. Word sense disambiguation provides a task-oriented evaluation of the quality of the acquired knowledge, but is not wholly satisfactory. We plan to explore the of use lexical resources such as (Magnini & Cavaglia, 2000) in the future.

Clean topic signatures open the avenue for interesting ontology enhancements, as they provide concepts with rich topical information. For instance, similarity between topic signatures could be used to cluster topically related concepts. Besides, word sense disambiguation methods could profit from these richer ontologies, and improve word sense disambiguation performance.